\documentclass{article}
\usepackage{spconf,amsmath,graphicx}
\usepackage{times}
\usepackage{epsfig}
\usepackage{graphicx}
\usepackage{amsmath}
\usepackage{amssymb}
\usepackage{amsfonts}
\usepackage{float}
\usepackage{array}

\usepackage{fancyhdr}
\usepackage{extramarks}

\newcommand{\tabincell}[2]{\begin{tabular}{@{}#1@{}}#2\end{tabular}}

\title{Learning Spatio-Temporal Representations \\ 
with Temporal Squeeze Pooling}
\name{Guoxi Huang and Adrian G. Bors}
\address{Department of Computer Science, University of York, York YO10 5GH, UK}

\begin{document}
%
\maketitle

\begin{abstract}
In this paper, we propose a new video representation learning method, named Temporal Squeeze (TS) pooling, which can extract the essential movement information from a long sequence of video frames and map it into a set of few images, named Squeezed Images. By embedding the Temporal Squeeze pooling as a layer into off-the-shelf Convolution Neural Networks (CNN), we design a new video classification model, named Temporal Squeeze Network (TeSNet). The resulting Squeezed Images contain the essential movement information from the video frames, corresponding to the optimization of the video classification task. We evaluate our architecture on two video classification benchmarks, and the results achieved are compared to the state-of-the-art.
\end{abstract}

\vspace*{-0.2cm}
\section{Introduction}
\label{sec:intro}
\vspace*{-0.2cm}

Video classification attracts increasing research interest, given its numerous applications. Extracting features characteristic to the movement in the scene is essential for applications from video surveillance to video summarization. We can identify two categories of approaches: hand-crafted feature representation methods, and deep learning-based feature methods.

Hand-crafted features, such as 3D histograms of gradients \cite{klaser2008spatio}, scale-invariant spatio-temporal interest points \cite{laptev2005space,willems2008efficient} dense trajectories \cite{wang2013action} as well as the dynamics of change in movement and location \cite{GroupAction2018}, have been used in various video tasks. Bilen {\em et al.} \cite{bilen2017action} proposed dynamic images that summarize spatio-temporal information of a video clip into a single image which preserves the semantics of the scene in a compact format.

Convolution Neural Networks (CNN) have been used to learn visual representations in many applications, and recently they have been employed successfully for video processing. The two-stream video classification model \cite{simonyan2014two} contains both spatial and temporal processing pipelines. CNNs models containing 3D convolution kernels (3D CNN), such as C3D \cite{tran2015learning} and I3D \cite{carreira2017quo}, represent a promising way for spatio-temporal representation learning. However, 3D CNNs are prone to overfitting when trained on small datasets from scratch. Meanwhile, the training of 3D CNN on large datasets requires great computational demands, and the model size is quadratic compared to 2D CNNs used for video processing. Long short-term memory (LSTM) networks \cite{hochreiter1997long}, represent a category of recursive neural networks (RNN) that can learn the long-term dependency of time series data and can compensate for the shortcomings of 2D CNNs to some extent. 2D CNN$+$LSTM \cite{yue2015beyond} was proposed to capture the spatio-temporal information from videos. However, 2D CNN$+$LSTM  was shown to have lower performance than the two-stream model in action recognition benchmarks, \cite{carreira2017quo}.

In this study, a new Temporal Squeeze (TS) pooling methodology, which can be embedded into CNNs, is proposed. The proposed TS pooling approach aggregates the temporal video information into a reduced spatial dimension by means of an optimization approach that preserves the video information characteristics. In this study, TS pooling is optimized for the video classification task. TS pooling can compensate for the shortcomings of the dynamic images \cite{bilen2017action}, by controlling the pooling size. This study demonstrates that the proposed TS pooling mechanism can summarize the visual representation of up to 64 video frames while dynamic images would only process 10 frames. By embedding the temporal squeeze pooling as a layer into the off-the-shelf CNNs, we design a new video classification model named Temporal Squeeze Network (TeSNet). The proposed methodology for representing video information is presented in Section~\ref{Sec3}. The experimental results are provided in Section~\ref{Sec4}, and the conclusion is drawn in 
Section~\ref{Conclu}.

\vspace*{-0.2cm}
\section{Video information representation}
\label{Sec3}
\vspace*{-0.2cm}


\subsection{Temporal Squeeze Pooling}
\vspace*{-0.1cm}

The proposed approach relies on the observation that consecutive video frames usually contain repeating information, especially either the background for a still camera, or the foreground, when the camera follows a target. A temporal squeeze pooling aims to compress the dynamic information of a video clip with \(K\) frames into \(D\) frames (\(D<<K\)), such that essential information is preserved. Consequently, repeating information is filtered out while preserving the essential, usually specific movement patterns. Let \(\mathbf{X}=[\mathbf{x}_1,\mathbf{x}_2,\ldots,\mathbf{x}_K] \) denote \(K\) video frames where \( \mathbf{x}_i \in \mathbb{R}^{H\times W \times C}\), $i=1,\ldots,K$ and \(H\), \(W\), \(C\) represent the height, width and the number of channels (colour), respectively. The TS layer aims to find out the optimal hyperplane \(\mathbf{A} \in \mathbb{R}^{K \times D}\), and map every pixel of \(\mathbf{X}\) from the vector space of $\mathbb{R}^{K}$ onto a much smaller information defining space $\mathbb{R}^{D}$. The aim is to preserve the relevant dynamic information across the temporal direction into the compressed space.

In the following, the squeeze and excitation operations proposed in \cite{Hu_2018_CVPR} are adopted for the TS pooling. The frame sequence $\mathbf{X}$ is initially processed by the squeeze operation, producing a frame descriptor. The squeeze operation is implemented by using global average pooling along the spatial dimensions \(H\), \(W\) and \(C\). Then, the squeeze operation is followed by the excitation operation, which is made up of two consecutive fully connected (FC) layers. The output of the excitation operation is reshaped to become the column space of $\mathbf{A}$, which defines a hyperplane. In the squeeze operation, the $k$-th element of a frame-wise  \(\mathbf{z} \in \mathbb{R}^K\) is calculated by:
\vspace*{-0.2cm}
\begin{equation} 
    z_k=F_{sq}(\mathbf{x}_k)=
    \frac{1}{H W C}\sum_{i=1}^H
    \sum_{j=1}^W\sum_{l=1}^Cx_k(i,j,l).
    \label{eq1}
    \vspace*{-0.2cm}
\end{equation}
In the excitation operation, the input-specific hyperplane is calculated by:
\vspace*{-0.2cm}
\begin{equation}
F_{ex}(\mathbf{z},\mathbf{W})=\delta_2(\mathbf{W}_2 \delta_1 (\mathbf{W}_1\mathbf{z})),
\label{eq2}
\vspace*{-0.2cm}
\end{equation}
where \(\delta_1\) and \(\delta_2\) refer to the activation functions and  \(\mathbf{W}_1 \in \mathbb{R}^{K \times K}\), \(\mathbf{W}_2 \in \mathbb{R}^{KD \times K} \) refer to the weights of the FC layers. Then, the output of equation (\ref{eq2}) is reshaped into a matrix \(\mathbf{A}' \in \mathbb{R}^{K \times D}\). The input-specific hyperplane for the projection is given by
\begin{equation} 
\mathbf{A} = \Phi(\mathbf{A}'),
\label{eq2_1}
\end{equation}
where $\Phi$ is a function that guarantees $\mathbf{A}$ is column independent. We flatten \(\mathbf{X}\) along its \(H\), \(W\) and \(C\) dimensions into a vector \(\overline{\mathbf{X}} = [\bar{\mathbf{x}}_1,\bar{\mathbf{x}}_2,...,\bar{\mathbf{x}}_{HWC}]\) where \(\bar{\mathbf{x}}_i \in \mathbb{R}^{K}\), and then project it onto the hyperplane $\mathbf{A}$, resulting in a vector \(\widehat{\mathbf{X}}\). The $i$-th element of the projection, \(\widehat{\mathbf{x}}\) is calculated by:
\begin{equation} \label{eq3}
\begin{aligned}
&\mathbf{y}_i = (\mathbf{A}^\mathsf{T} \mathbf{A})^{-1} \mathbf{A}^\mathsf{T} \bar{\mathbf{x}}_i, \\
&\widehat{\mathbf{x}}_i = \mathbf{A} \mathbf{y}_i, \quad  i=1,\ldots,HWC
\end{aligned}
\end{equation}
where \(\mathbf{y}_i \in \mathbb{R}^D \) represent the mapping coefficients. We reshape the vector \(\mathbf{Y} =  [\mathbf{y}_1,\mathbf{y}_2,\ldots,\mathbf{y}_{HWC}] \) into a new image sequence \(\mathbf{Y}'\) of $D$ frames of size $H \times W \times C$. The squeezed sequence of $D$ frames can be used as a simplified, yet an information comprehensive representation, that summarizes the dynamics taking place in the given set of $K$ video frames.

\vspace*{-0.1cm}
\subsection{Optimization}
\vspace*{-0.1cm}

In this study, the TS pooling is optimized with respect to the video classification task. In order to ensure that the projection \(\widehat{\mathbf{X}}\) retains as much meaningful spatio-temporal information as possible from the original  video sequence \(\mathbf{X}\), \(\widehat{\mathbf{X}}\) should be close to the original video data \(\overline{\mathbf{X}}\). This relies on finding the optimal hyperplane \(\mathbf{A}\) fitting \(\mathbf{X}\) aiming to minimize the residuals of projections. Let us denote the mean absolute error (MAE) on projections by $l_{proj}$, calculated as:
\vspace*{-0.2cm}
\begin{equation} 
l_{proj} = \frac{1}{H W C}\sum_{i}^{HWC} \left \| \bar{\mathbf{x}}_i - \widehat{\mathbf{x}}_i \right \|,
\label{eq4}
\vspace*{-0.2cm}
\end{equation}
where $\left \|\cdot  \right \|$ represents the standard $L^2$ norm in the K-dimensional Euclidean space $\mathbb{R}^K$.

\vspace*{-0.2cm}
\subsection{Temporal Squeeze Network} 
\label{TempSquNet}
\vspace*{-0.1cm}

The Temporal Squeeze pooling can process not just video frames but also the outputs of convolution layers of a CNN. When it is plugged into off-the-shelf CNNs, it forms a new architecture, named Temporal Squeeze Network (TeSNet). We choose the Inception-ResNet-V2 \cite{szegedy2017inception} as the backbone CNN embedding the TS pooling block. In order to form an end-to-end training, we add the loss term \(l_{proj}\) from (\ref{eq4}) to the classification loss used in the original network, resulting in the following loss function:
\vspace*{-0.2cm}
\begin{equation}
l_{final} = l_{classif} + \beta \sum_{i=1}^{M} l_{proj}^i + \lambda l_{L2},
\label{FinalLoss}
\vspace*{-0.2cm}
\end{equation}
where \(l_{classif}\) is the cross-entropy loss of the classification \cite{szegedy2017inception}, \(l_{L2}\) is the L2 normalization term of all the trainable weights in the architecture, \(\lambda\) is the weight decay, \(\beta\) is the weight for the TS loss component \(l_{proj}\), where the projection residuals are summed up for all $M$ TS layers.

TS layers can be embedded in different sections of the backbone CNN. We design our model by following the principle of decreasing the number of mapped frames \(D\) when embedding into a deeper network layer position. In this case, the model represents a pyramidal video processing scheme. The first TS layer should be configured with a relatively larger \(D\) generating more frames, and therefore the loss of temporal information caused by successive pyramidal projections would be reduced.
We adopt the two-stream architecture \cite{simonyan2014two}, including an RGB image frame stream and an Optical Flow (OF) stream. For the OF stream, we use the TV-L1 optical flow algorithm \cite{zach2007duality} and its output is stored as JPEG images, where the colour encodes the optical flow vectors. 

\begin{figure*}[h]
\begin{center}
\begin{tabular}{cc}
\includegraphics[width=.75\textwidth]{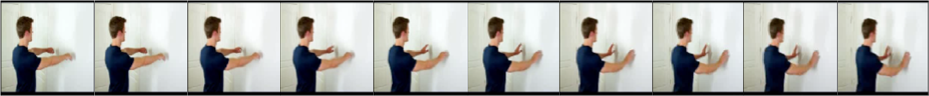}& \includegraphics[width=.15\textwidth]{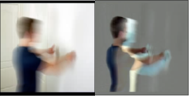}\\
\includegraphics[width=.75\textwidth]{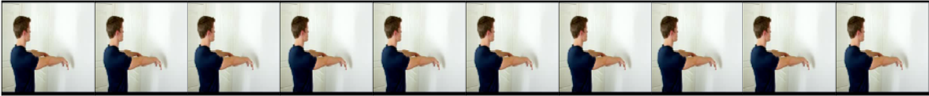}& \includegraphics[width=.15\textwidth]{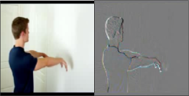}\\
(a) {\small Selection of 10 consecutive frames from video sequence.}  & (b) {\small Output of TS} \\
\end{tabular}
\end{center}
\vspace*{-0.6cm}
\caption{\label{fig:motion_evaluation} Visualizing the input and the corresponding output of the TS layer with $K=10$, $D=2$.}
\vspace*{-0.4cm}
\end{figure*}

\vspace*{-0.2cm}
\section{Experimental results}
\label{Sec4} 
 \vspace*{-0.2cm}

\subsection{Dataset and Implementation Details} \label{Datasets and Implementation Details}

We conduct experiments on two human activity classification benchmarks, UCF101 \cite{soomro2012ucf101} and HMDB51 \cite{kuehne2011hmdb}. UCF101 contains 13,320 real video sequences labelled int 101 classes, collected from YouTube \cite{soomro2012ucf101}, while HMDB51 contains 7,000 video clips distributed in 51 action classes \cite{kuehne2011hmdb}.
Our model is pre-trained on ImageNet \cite{imagenet_cvpr09}. To evaluate our model, we reimplement the Temporal Segment Network (TSN) \cite{wang2016temporal} with our backbone network. We set the dropout as 0.5 to prevent overfitting and adopt the same data augmentation techniques as in \cite{wang2016temporal} for network training. The size of the input frames is set to $299 \times 299$, which is randomly cropped from the resized images, and $K$ consecutive frames are randomly selected from each video sequence. We use Stochastic Gradient Descent for optimizing the network parameters in which the batch size is set to 32, momentum of 0.9, weight decay $\lambda = 4e^{-5}$, $\beta=10$. The initial learning rate is set to 0.001 for the image stream and at 0.005 for the Optical Flow stream. We train the model for 30 epochs, with a ten times reduction for the learning rate when the validation accuracy saturates. 

During testing, we uniformly sample 20 clips from each video clip and perform spatially fully convolutional inference for all clips, and the video-level score is obtained by averaging all the clip prediction scores of a video. For the proposed TeSNet, we set $\Phi(\cdot)=I$ in (\ref{eq2_1}), resulting in $\mathbf{A}'=\mathbf{A}$, while the column independent $\mathbf{A}$ is properly initialized. 
 We consider LeakyReLU for $\delta_2(\cdot)$ and the Sigmoid activation function for $\delta_1(\cdot)$ in equation (\ref{eq2}) and these choices are crucial for the performance of the model.

\vspace*{-0.2cm}
\subsection{Visualization Analysis}
\vspace*{-0.1cm}

 We explore how the temporal squeeze pooling represents the spatio-temporal information within the video clips by visualizing its outputs. In Fig~\ref{fig:motion_evaluation}, we show the output of the TS layer with $K=10$, $D=2$ resulting in 2 squeezed images. The clip, shown on the first row in Fig~\ref{fig:motion_evaluation}a display a clear salient movement, and we can observe that its corresponding output of TS summarizes the spatio-temporal information, as shown on the first row in Fig~\ref{fig:motion_evaluation}b. The other clip, shown on the second row, does not contain any obvious movement. When there is no movement present in a video clip, the TS layer captures the characteristic static information about the scene, as shown in the last two images from the second row of Fig~\ref{fig:motion_evaluation}b.

\begin{figure}[ht]
\vspace*{-0.2cm}
\begin{center}
\begin{tabular}{cc}
\includegraphics[width=.15\textwidth]{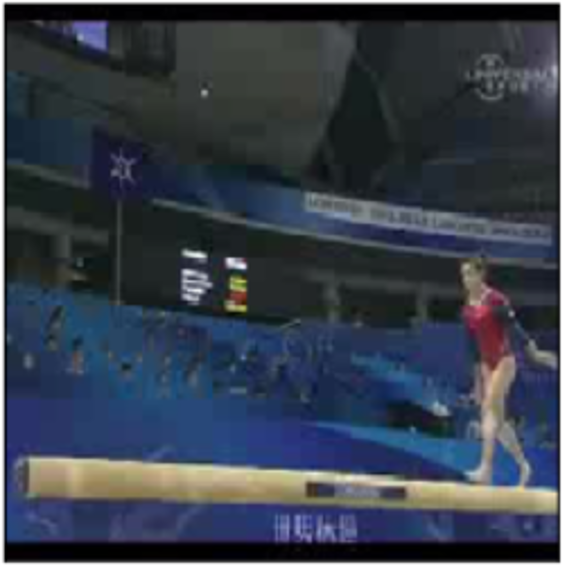}& \includegraphics[width=.30\textwidth]{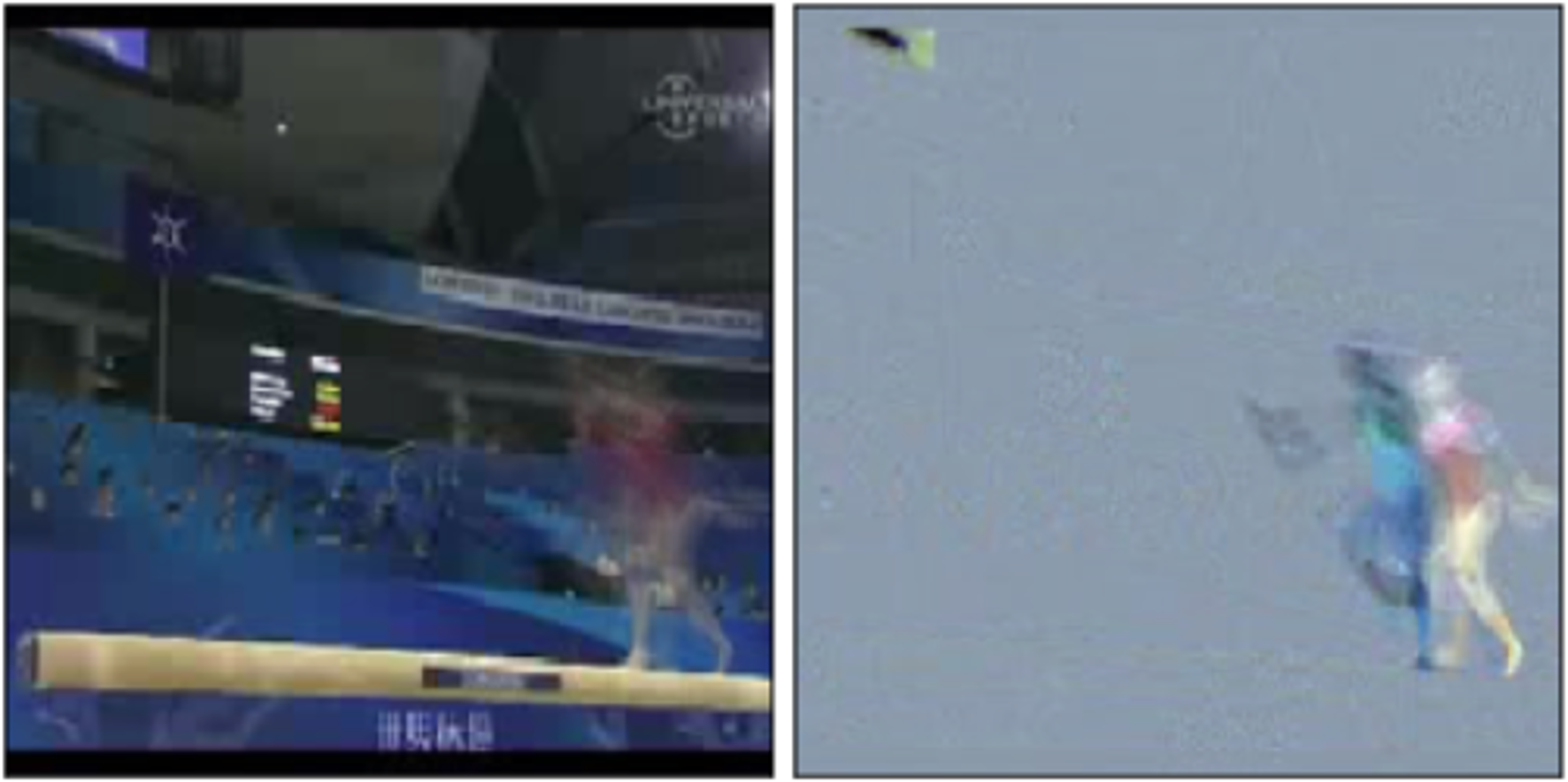}\\
(a) {\small Single Image}  & (b) {\small Squeezed Image} \\
\includegraphics[width=.15\textwidth]{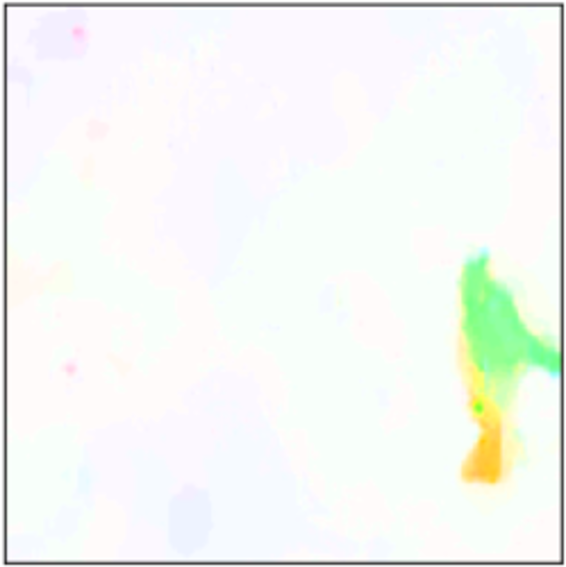}& \includegraphics[width=.30\textwidth]{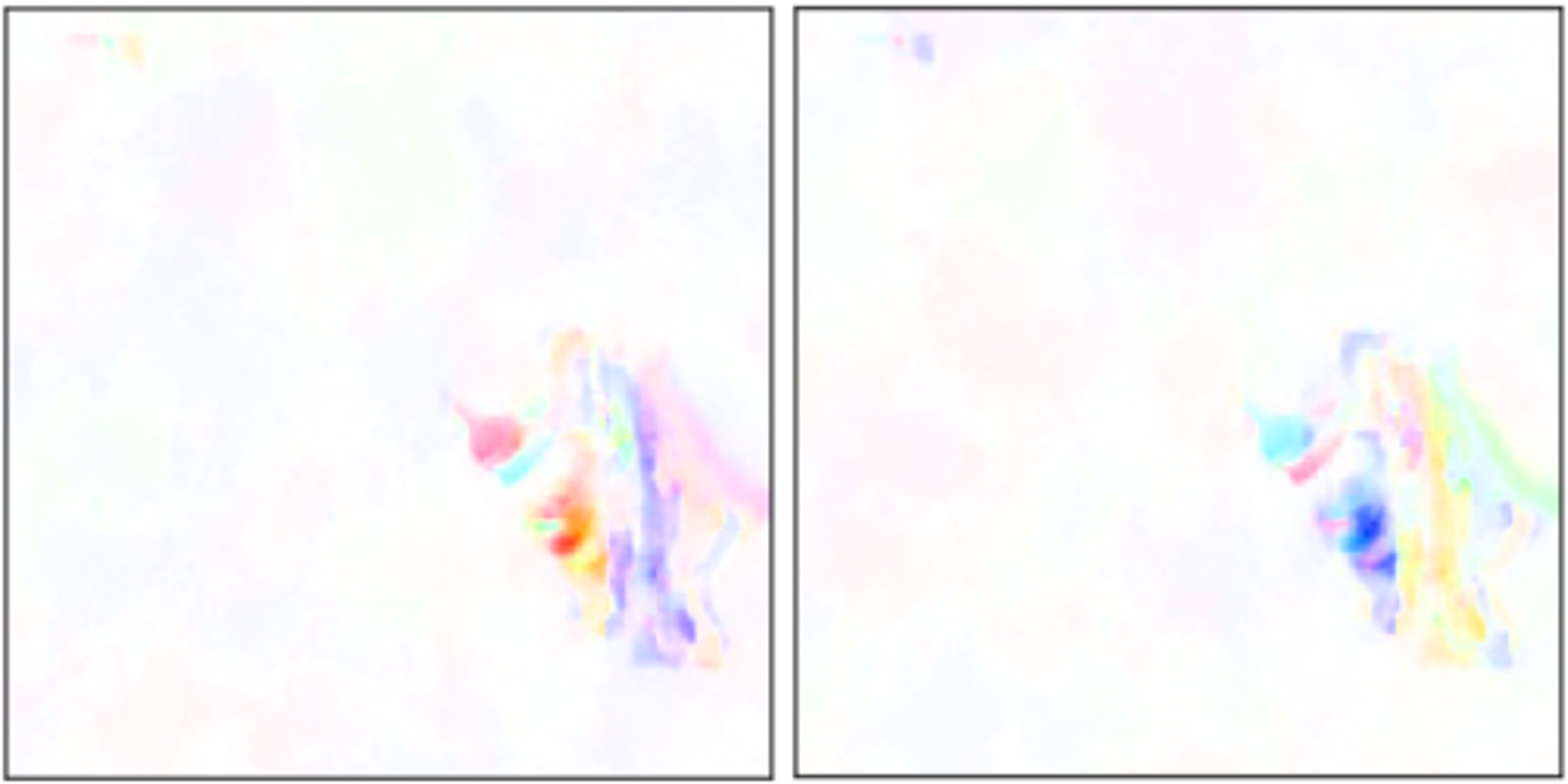}\\
(c) {\small Optical Flow} & (d) {\small Squeezed Optical Flow} \\
\end{tabular}
\vspace*{-0.4cm}
\end{center}
\caption{Given input video frames, flow images and the corresponding outputs for the TS layers $K=10$, $D=2$.}
\label{fig:four_output}
\vspace*{-0.2cm}
\end{figure}

Fig~\ref{fig:four_output} depicts the outputs of the TS layer with $K=10$, $D=2$. The output of the TS layer with RGB frames is shown in Fig.~\ref{fig:four_output}b, and the output of the TS layer of optical flow images is shown in Fig.~\ref{fig:four_output}d. We observe that the output of the TS layer tends to preserve the still information and the motion information separately. This indicates that by considering a single image we may not be able to represent the underlying spatio-temporal information from the video. Moreover, when considering $D=3$, the classification accuracy is higher than 
for $D=1$, according to the results from Table \ref{table: position_exploration}. This result further demonstrates that summarizing the dynamics of a long video clip into a single image would lose essential spatio-temporal information. A dynamic image \cite{bilen2017action} attempts to summarize the entire information from a video clip into a single image, which can explain why they fail to represent long video clips properly.

\vspace*{-0.2cm}
\subsection{Embedding the TS layer into the network}
\vspace*{-0.1cm}

In the following, we explore where and how to embed TS layers into the CNN. The results are shown in Table~\ref{table: position_exploration}, where the second column indicates the location for inserting a TS layer with the corresponding $D$ indicated in the third column. A single TS layer, $M=1$ is embedded in settings No. 1 and 2, while $M=2$ for settings No. 3, 4 and 5.
The model from the No. 1 setting, which embeds a TS layer directly after the network's inputs, achieves the best result in all settings. However, the model with the No. 5 setting, which embeds two TS layers into the backbone network, requires less computation and has almost the same performance  as the No. 1 setting. When a lower level of computational complexity is required, then the No. 5 setting is preferable to be used. Inserting the TS layers into the middle section of the backbone network leads to worse performance. One possible explanation is that the network was initially pretrained on ImageNet; the inserted TS layers did not fit well with the settings of these pretrained kernels and resulted in poor performance. In order to avoid this problem, the model with TS layers embedded needs pretraining on a large video dataset.
\begin{table}[ht]
\begin{center}
\begin{tabular}{c|l|c|c}
\hline
No.  & Location of & Number of squeezed   & Top-1 \\ 
  & TS layer    &    frames ($D_i$) &   (\%)   \\
\hline
\hline
1 & Input  & $D_1=3$ & 85.4 \\
\hline
2 & Input  & $D_1=1$ & 83.1 \\
\hline
3 & \tabincell{l}{Conv2d\_1a\_3x3\\ Conv2d\_4a\_3x3 }  & \tabincell{l}{ $D_1=3$\\ $D_2=1$ } & 81.7\\
\hline
4 & \tabincell{l}{Conv2d\_1a\_3x3\\ Block A }  & \tabincell{l}{ $D_1=3$\\ $D_2=1$ } & 84.9\\
\hline
5 & \tabincell{l}{Conv2d\_1a\_3x3\\ Block B }  & \tabincell{l}{  $D_1=3$ \\ $D_2=1$ } & 85.3 \\
\hline
\end{tabular}
\end{center}
\vspace*{-0.2cm}
\caption{Evaluating the accuracy when embedding the TS layer at different depths of the CNN.}
\label{table: position_exploration}
\vspace*{-0.2cm}
\end{table}

We explore how the clip length of videos affects performance. Here, we use a rather small batch size of 8 because of the GPU memory limitation. For a clip length of 64, we adopt setting No. 5 from Table \ref{table: position_exploration} but consider $D_1=16$ and $D_2=4$. When considering clip lengths of 10 or 16, we use the first setting from Table \ref{table: position_exploration}. The results are shown in Table \ref{table: clip_length}. Due to the small batch size we adopt, we see little performance degradation for the model with a clip length of 10. Nevertheless, it can be observed that when increasing the length of the video clip, the performance improves as well.
\begin{table}[ht]
\begin{center}
\begin{tabular}{c|c}
\hline
clip length &  Classif. (\%) \\ 
\hline
\hline
10 & 85.3\\
16 & 86.2\\
64 & 87.8\\
\hline
\end{tabular}
\end{center}
\vspace*{-0.2cm}
\caption{Comparing the effect of various clip length of videos on RGB stream on the split 1 of UCF101 database.}
\vspace*{-0.2cm}
\label{table: clip_length}
\end{table}
\begin{table}[ht]
\begin{center}
\begin{tabular}{l|c|c|c|c}
\hline
Architecture & length & RGB & OF & RGB + OF \\
\hline
\hline
Baseline  & 1 & 83.5 & 85.4 & 92.5 \\
TSN  & 3 & 85.0 & 85.1 & 92.9\\
TeSNet  & 64 & 87.8 & 88.2 & 95.2 \\
\hline 
\end{tabular}
\end{center}
\vspace*{-0.2cm}
\caption{Performance of different architectures with two-stream on the split 1 of UCF101 database.}
\label{table: two-stream acc} 
\vspace*{-0.2cm}
\end{table}

In order to evaluate the effectiveness of the proposed TeSNet architecture, we compare it with the baseline and TSN~\cite{wang2016temporal}. All these models adopt Inception-ResNet-v2 as their backbone networks. The results provided by different architectures and streams are shown in Table \ref{table: two-stream acc}. After employing the proposed TeSNet architecture with the fusion of the RGB and optical flow (OF) modalities, we successfully boosted the Top-1 accuracy from 92.5\% to 95.2\% on the split 1 of UCF101. TeSNet also outperforms TSN (Inception-ResNet-v2) by 2.3\%, which strongly demonstrates the effectiveness of TeSNet.

\begin{table}[ht]
\centering
\begin{tabular}{l|c|c}
\hline
Method &UCF101 & HMDB51\\
\hline
\hline 
iDT+Fisher vector \cite{peng2014action} & 84.8 & 57.2 \\
 iDT+HSV \cite{peng2016bag}& 87.9 & 61.1 \\
C3D+iDT+SVM \cite{tran2015learning}  & 90.4 & - \\
\tabincell{l}{ Two-Stream (fusion by SVM) \cite{simonyan2014two}}  & 88.0 & 59.4 \\
Two-Stream Fusion+iDT \cite{feichtenhofer2016convolutional} & 93.5 & 69.2  \\
\tabincell{l}{ TSN (BN-Inception) \cite{wang2016temporal}} & 94.2 & 69.4  \\
Two-Stream I3D \cite{carreira2017quo}&93.4 & 66.4\\
TDD+iDT \cite{wang2015action} & 91.5 & 65.9 \\
Dynamic Image Network \cite{bilen2017action}  & 95.5 & 72.5\\
\hline
Temporal Squeeze Network  & 95.2 & 71.5 \\
\hline
\end{tabular}\\ 
\caption{Temporal squeeze network compared with other methods on UCF101 and HMDB51, in terms of top-1 accuracy, averaged over three splits.} 
\label{T4}
\vspace*{-0.2cm}
\end{table}

\subsection{ Comparison with the state-of-the-art}
\label{sec:comparison}

For fair comparisons, we only consider those models that are pre-trained on ImageNet \cite{imagenet_cvpr09}. The results are provided in Table~\ref{T4}. The proposed TeSNet achieves 95.2\% top-1 accuracy on UCF101 and 71.5\% on HMDB51, which outperforms TSN (BN-Inception) by 1\% and 2.1\% on UCF101 and HMDB51, respectively. The dynamic image network fuses the prediction scores of four streams using a better backbone network architecture, while the proposed model only uses two streams, so the results are not directly comparable. Nevertheless, the advantage of our proposed method is that we can control the number of frames for the output of the TS layer, while the dynamic image method \cite{bilen2017action} can only summarize a part of the spatio-temporal information into a single image. The proposed TeSNet method can represent the information through TS pooling from as many as 64 frames, unlike in \cite{bilen2017action}, where the dynamic image method would show performance degradation when processing more than 20 frames.

\section{Conclusion} 
\label{Conclu}

This paper proposes a new video representation scheme while aiming to improve the models on video classification tasks, called Temporal Squeeze (TS) pooling. By embedding the TSP module into off-the-shelf CNNs, we proposed the Temporal Squeeze Network (TeSNet) architecture, which learns spatio-temporal features characteristic to discriminating classes of video sequences. We have investigated various locations in the structure of the CNN network for embedding the TSP layers. TeSNet is also used on the optical flow data stream, the perdition score of which is then fused with the RGB image stream, which produces higher accuracy than any single stream on video classification tasks. Experiments have been performed on both UCF101 and HMDB51 datasets, and the results indicate that the new video representations are compact and meaningful for improving the video classification performance.

\vfill\pagebreak
\bibliographystyle{TempSqueezeICASSP20.bbl}
\bibliography{strings,refs}

\end{document}